\documentclass{article}
\usepackage{spconf,amsmath,epsfig}
\usepackage{graphicx}
\usepackage{subfigure}
\usepackage{colortbl}
\usepackage[table*]{xcolor}
\usepackage{algorithm}
\usepackage{algorithmic}
\usepackage{multirow}
\usepackage{multicol}
\usepackage{amssymb}
\usepackage{amsmath}
\usepackage{cite}
\usepackage{caption}
\usepackage{multirow}
\usepackage{booktabs}

\title{Pixel DAG-Recurrent Neural Network for
Spectral-Spatial Hyperspectral Image Classification}
\name{Xiufang Li, Qigong Sun, Lingling Li, Zhongle Ren, Fang Liu, Licheng Jiao
\thanks{This work was supported in part by the State Key Program of National Natural Science of China (No.61836009, No. 91438201 and No. 91438103),
the National Natural Science Foundation of China (No. 61871310, No. 61876220, No. 61801351)).}}

\address{Key Laboratory of Intelligent Perception and Image Understanding of Ministry of Education,\\
International Research Center for Intelligent Perception and Computation,\\
Joint International Research Laboratory of Intelligent Perception and Computation,\\
School of Artificial Intelligence, Xidian University, Xi’an, Shaanxi Province 710071, China}
\begin{document}
%\ninept
%
\maketitle
\begin{abstract}
Exploiting rich spatial and spectral features contributes to improve the classification accuracy of hyperspectral images (HSIs).
In this paper, based on the mechanism of the population receptive field (pRF) in human visual cortex, we further utilize the spatial correlation of pixels in images and propose pixel directed acyclic graph recurrent
neural network (Pixel DAG-RNN) to extract and apply spectral-spatial
features for HSIs classification. In our model, an undirected
cyclic graph (UCG) is used to represent the relevance connectivity of pixels
in an image patch, and four DAGs are used to approximate
the spatial relationship of UCGs. In order to avoid overfitting, weight sharing and dropout are adopted. The higher classification performance of our model on HSIs classification has been verified by experiments on three benchmark data sets.
\end{abstract}
%\vspace{-0.1cm}
\section{Introduction}
%\vspace{-0.2cm}
\label{sec:intro}

HSIs usually provide abundant spectral and spatial information of ground targets. Therefore, their interpretations such as
the classification has been widely used in the geological survey, vegetation research and so on.
However, the sufficient and efficient utilization of spectral and spatial information in HSIs classification is challenging such as Hughes phenomenon\cite{Plaza2009Recent}. A large number of researches have been done on the classification of HSIs from traditional methods such as independent component analysis (ICA)\cite{villa2011hyperspectral} to popular deep convolutional neural network (DCNN)\cite{yue2015spectral}.

Due to the increase of spectral dimension and nonlinear of spectral space, some traditional methods are not appropriate to classify HSIs. Subsequently, ensemble learning method is proposed for HSIs classification such as \cite{waske2010sensitivity}.
However, most of ensemble learning methods only consider the spectral rather than spectral-spatial information.
Then, some sparsity-based algorithms, for instance \cite{Wang2015Hyperspectral}, are used to extract spatial-spectral features. These methods are incapable of capturing robust and abstract features with the complex and varied environment.

Recently, deep neural network (DNN) has attracted more attentions due to its advantages of extracting high-rank abstract features and achievements in computer vision and natural language process, such as the images classification\cite{krizhevsky2012imagenet} and speech recognition \cite{graves2013speech}. Y. Chen \textsl{et al}.\cite{chen2016deep} proposed 3-D CNN, it is the first time to extract spectral-spatial features of HSIs simultaneously by DCNN, and this method improves classification performance obviously.
L. Mou \textsl{et al}. \cite{Mou2017Deep} take hyperspectral pixels as sequential data and apply recurrent neural network (RNN) to classify HSIs.
%Further, a image density model based on pixel distribution is proposed and applied to predict target pixel by DRNN \cite{oord2016pixel}.
%Besides, Bing Shuai \textsl{et al}. also proposes a novel model based on spatial structure sequentiality to represent the image in order
These above methods extract robust spatial-spectral information
in HSIs classification by multilayer convolutional neural network. We know, these convolution processes used to extract pixel features utilize the ideology of receptive field. However, this operation only extracts the feature of every pixel and ignore the correlation of adjacent pixels which is significant to realize category in the pRF.
For detail, the perception ability of the pRF in the human visual cortex\cite{wandell2015computational} is related to the focus of the vision and surrounding scenes. The influence of surrounding scenes on the central target reduces with the increase of interval. Here, we use UCG and DAG to describe this relation. Due to the ability of RNN on dealing with sequence data, we proposed a Pixel DAG-RNN model for HSIs classification.

This new method has three advantages: (1) Our model further applies the pRF mechanism of the human visual system in identifying the pixel category. Besides the pixel feature, we further consider the spatial correlation of adjacent pixels. (2) We apply UCGs and DAGs to represent the correlation of pixels and RNN to apply spatial sequence features for HSIs classification. This model makes full use of both spectral information and spatial correlations of pixels.
(3) In order to prevent the overfitting phenomenon caused by a limited number of training samples, weight sharing and dropout are used.
\section{Pixel DAG-Recurrent Neural Network}
\label{sec:method}
\vspace{-0.2cm}
\subsection{Motivation}
\vspace{-0.3cm}
\begin{figure}[!htb]
\centering
\setlength{\abovecaptionskip}{-0.0cm}
\setlength{\belowcaptionskip}{-0.4cm}
\includegraphics[width=3.6in,height=0.9in]{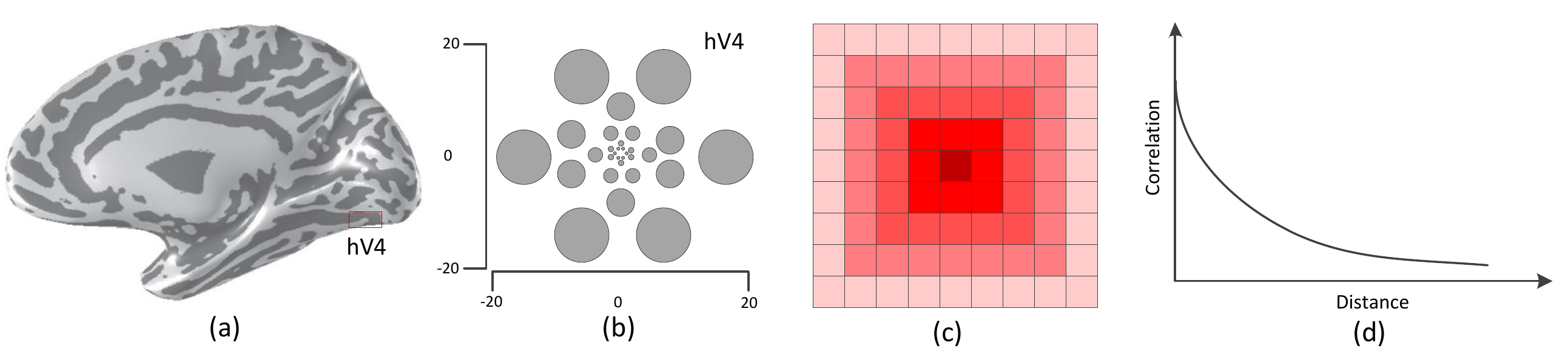}
\caption{\footnotesize{(a) Schematic diagram of the human cerebral cortex, where hV4 is marked by red rectangle box. (b) The pRFs spatial array of hV4. A series of different size circles are used to denote pRFs. (c) An $9 \times 9$ image patch with the center pixel represents target unit. (d) The diagram of correlation with distance}}
\label{fig_University}
\end{figure}

We know, in the human vision system, that sounding scenes also affect the realization of the central target and these effects obey visual mechanism. In the pRF mechanism of the human visual system, which illustrated in Fig.1 (a), more intensive attentions are focused on the central target and the attentions on sounding scenes because blur and sparse with increased distance as shown in Fig.1 (b). Based on this theoretical mechanism, we utilize the same principle of identifying pixel categories. The importance of surrounding layer pixels on central pixel reduces with the increased interval.
Fig.1 (c) shows the reduced importance of surrounding pixel layers on a central pixel by shallower color. Fig.1 (d) express the variance curve of the importance of surrounding pixel layers on the central pixel with the distance.
These rules demonstrate the importance of spatial structure sequentiality in understanding images.
In addition, RNN is more suitable for processing sequence data.
Therefore, on the basis of the mechanism of pRF mechanism of the human visual system in HSIs classification,
we used UCGs and DAGs to connect pixels and design Pixel DAG-RNN to extract spectral-spatial features.
\vspace{-0.3cm}
\subsection{Pixel DAG-RNN}
\vspace{-0.3cm}
\begin{figure}[!htb]
\centering
\setlength{\abovecaptionskip}{-0.0cm}
\setlength{\belowcaptionskip}{-0.4cm}
\includegraphics[width=3.0in,height=0.9in]{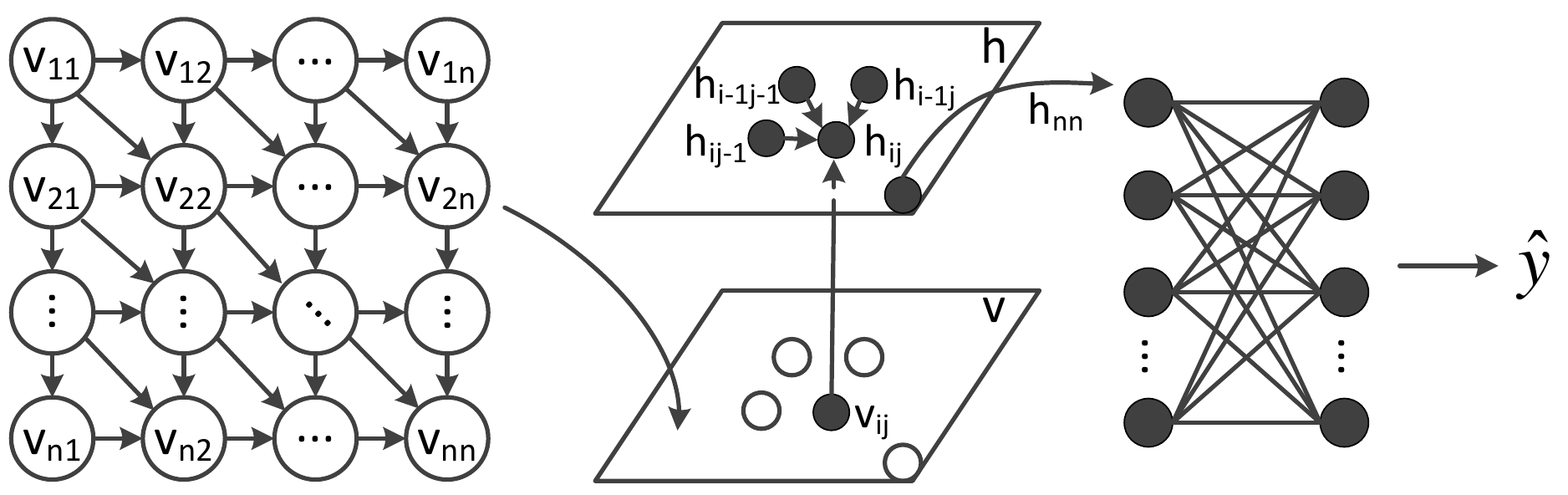}
\caption{\footnotesize{Architecture of Pixel DAG-RNN for classification. The leftmost is a sample of 8-neighborhood DAG in the southeast direction.}}
\label{fig_University}
\end{figure}
As shown in Fig. 2, a sample of directed acyclic graph (DAG) can be represented as $\mathcal{G} =\left\{ \mathcal{V},\mathcal{A} \right\}$,
where $\mathcal{V}=\left \{ v_{ij} \right \}$ denotes the vertex set and $\mathcal{A}=\left \{ \overrightarrow{v_{ij}v_{pq}} \right \}$ denotes the arc set, which $i, p$ represent the row and $j, q$ represent the column from $1$ to $n$, $\overrightarrow{v_{ij}v_{pq}}$ represents the arc from vertex $v_{ij}$ to $v_{pq}$. We can get a series of contextual dependencies among vertexes and input them to a recurrent neural network. Subsequently, hidden layer $h$ are generated with the same structure as $\mathcal{G}$. $h^{(v_{ij})}$ denotes the value of hidden layer at $v_{ij}$, which is related to its local input $x^{(v_{ij})}$ and the hidden representation of its predecessors. Due to the special structure of hidden layer $h$, we should calculate it sequentially.
The hidden layer $h$ and output $\hat{y}$ are computed as follows:
\begin{eqnarray}
\hat{h}^{(v_{ij})}&=&\sum_{v_{pq} \in \mathcal{P}_{{G}}(v_{ij})} {G}(Sh^{(v_{pq})}+a) \\
h^{(v_{ij})}&=&{H}(Ux^{(v_{ij})}+V\hat{h}^{(v_{ij})}+b) \\
\hat{y}&=&{F}(Wh^{(v_{nn})}+c)
\end{eqnarray}
where $S, U, V, W$ and $a, b , c$ represent the connection weights and biases. $\mathcal{P}_{\mathcal{G}}(v_{ij})$ is the direct predecessor set of vertex $v_{ij}$ in the DAG, ${F}(\cdot)$, ${G}(\cdot)$ and ${H}(\cdot)$ are the nonlinear activation functions. From the above formulas, we can see that this is an autoregressive model with the following conditional distributions:
\begin{eqnarray}
p(h^{(v_{ij})}|h^{(v_{i-1j-1})},h^{(v_{i-1j})},h^{(v_{ij-1})},x^{(v_{ij})})
\end{eqnarray}
The recurrent weights $S, U, V$ are shared in order to avoid overfitting.
When calculating the hidden layer $h$, we start at the DAG's source vertex $v_{11}$ and calculate the next vertex of the hidden layer according to the structure of DAG, until to the last vertex $v_{nn}$. Therefore, $h^{(v_{nn})}$ include the information of all the DAG's vertex. A nonlinear function ${F}(\cdot)$ is used to obtain the final output $\hat{y}$. Loss $l$ is denoted as follows:
\begin{eqnarray}
l=L(\hat y, y)
\end{eqnarray}
where $y$ represents real label, ${L}(\cdot)$ denotes loss function.
\subsection{Pixel DAG-RNN for HSIs Classification}
In HSIs classification, each pixel having hundreds of spectral data is classified into a class of object. In order to apply the neighborhood information, we use UCG to represent the spatial relationship of the image

\begin{figure*}[!htb]
\centering
\setlength{\abovecaptionskip}{-0.0cm}
\setlength{\belowcaptionskip}{-0.3cm}
\includegraphics[width=6.6in,height=2.0in]{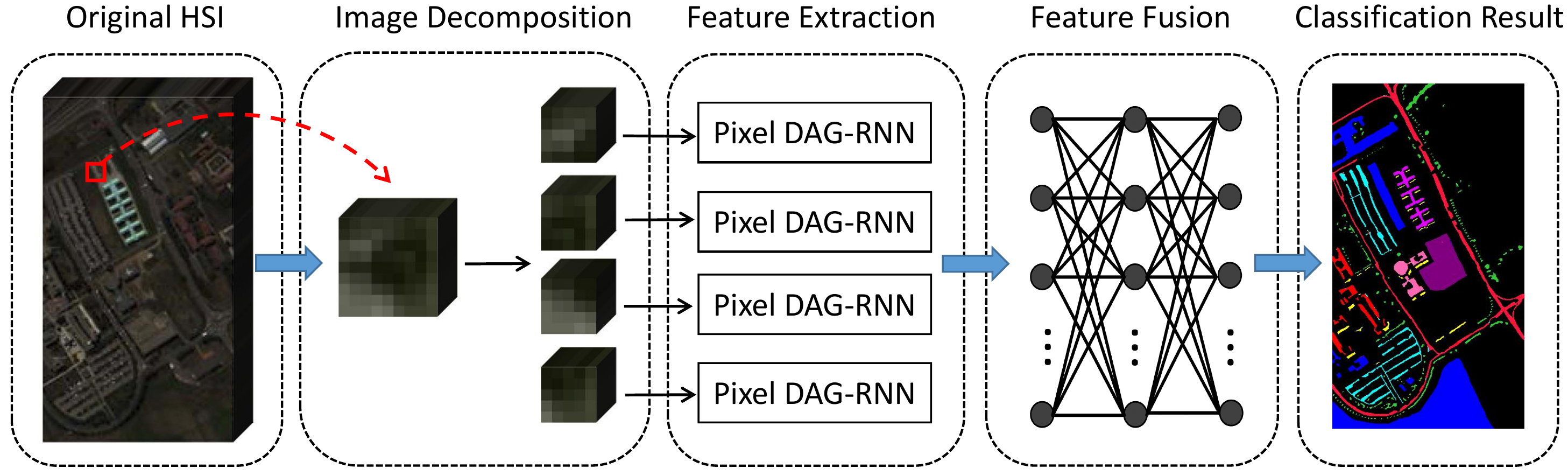}
\caption{\footnotesize{Architecture of Pixel DAG-RNN for HSIs classification}}
\label{fig_University}
\end{figure*}

patch and then apply four different DAGs to approximate the
topology of the UCG.
Based on DAG's definition in spatial structural sequentiality of pixels, we apply DAG-RNNs model to classify HSIs for making full use of the spatial structural sequentiality of pixels as shown in Fig. 3. The spectral-spatial
features are extracted by Pixel DAG-RNNs, and then we
concatenate the four feature vectors at the end vertex to a final vector. Two full connection layers and softmax are used for the final classification.

%\begin{figure}[!htb]
%\centering
%\setlength{\abovecaptionskip}{-0.0cm}
%\setlength{\belowcaptionskip}{-0.4cm}
%\includegraphics[width=2.5in,height=1.2in]{ch/Decomposition_8UCG_to_DAG.pdf}
%\caption{\footnotesize{Schematic diagram of image patch decomposition. The black unit denotes target pixel.}}
%\label{fig_University}
%\end{figure}
The detail of using DAGs to approximate UCG is illustrated in Fig. 4. Suppose we make an image patch with the size of $n \times n$. The black unit denotes the target pixel. 8 neighborhood UCG is used to represent the spatial relationship of pixels. Because of its loopy property, we can't get a fixed sequence applied in RNNs. In order to fully use the semantic contextual dependencies of the image patch, we use the combination of a set of small DAGs with the height and width being ($n/2+1$) to represent the UCG. Four 8 neighborhood DAGs are respectively used in the end vertex as the target unit.
Those four dictionaries are southeast, southwest, northeast, and northwest. Therefore, we can route anywhere orderly and use the information of any pixel in the image patch. The order of calculation is row by row and pixel by pixel within every row, suggested by \cite{van2016conditional}. Pixel DAG-RNN is applied to each DAG to generate the hidden layer $h_d$ ($d\in \left \{ southeast, southwest, northeast, northwest \right \}$), so as to take advantage of the local feature with a broader view of contextual awareness.
Those operations can be expressed as follows:
\begin{equation}
\hat{h}^{(v_{ij})}_d=\sum_{v_{pq} \in \mathcal{P}_{\mathcal{G}_d}(v_{ij})} {G}(Sh^{(v_{pq})}+a)
\end{equation}
\begin{equation}
h^{(v_{ij})}_d={H}(Ux^{(v_{ij})}+V\hat{h}^{(v_{ij})}_d+b)
\end{equation}
where $S,U, V$ and $a, b$ are the connection weights and biases, $h^{(v_{ij})}_d$ is the hidden layer at vertex $v_{ij}$ in d direction DAG.
$\mathcal{P}_{\mathcal{G}_d}(v_{ij})$ is the direct predecessor set of vertex $v_{ij}$ in direction $d$                        .
Here, the weights and biases are shared across all vertexes in direction $d$. The memory length is ($n/2+1$).
\begin{figure}[!htb]
\centering
\setlength{\abovecaptionskip}{-0.0cm}
\setlength{\belowcaptionskip}{-0.4cm}
\includegraphics[width=2.6in,height=1.2in]{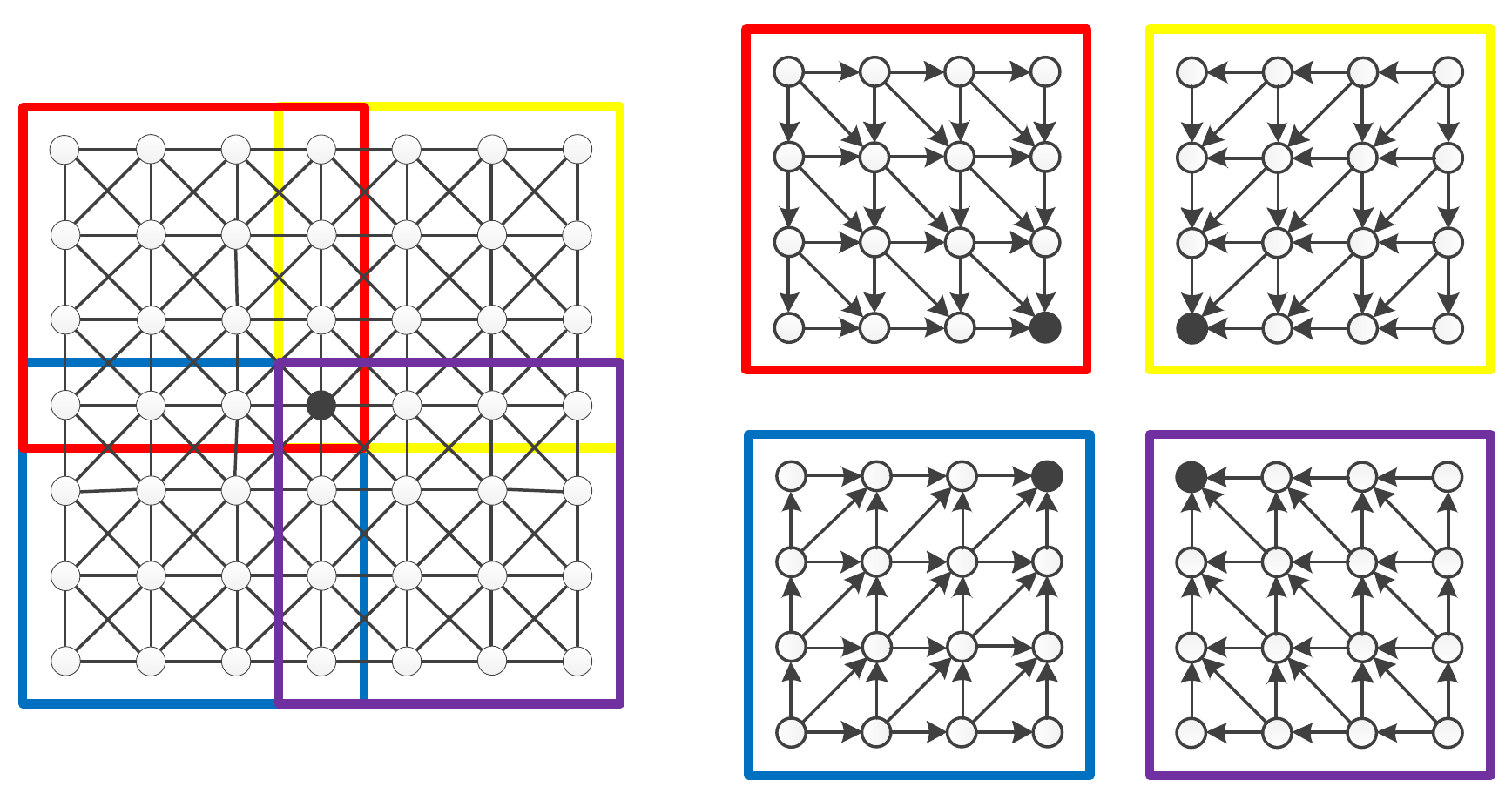}
\caption{\footnotesize{Schematic diagram of image patch decomposition. The black unit denotes target pixel.}}
\label{fig_University}
\end{figure}
Our proposed model can sequentially get the feature from edge to center, because of the accumulation of the parameters in the forward process, the information of the neighborhood is gradually weakened. It is consistent with the principle: the importance of surrounding pixel to central pixel reduces with the increased interval.
\section{EXPERIMENTAL RESULTS AND ANALYSIS}
\label{sec:experiment}
In this experiment, we use three benchmark datasets: Indian
Pines data, University of Pavia data, and Kennedy Space Center data and their size are $145\times145$, $610\times340$, and $512\times614$, respectively. Their usable number of bands are 200, 115 and 176 respectively. In addition, in order to verify the effectiveness of our method, we choose the RBF-SVM, SOMP, 3D-CNN as contrast experiments and take overall accuracy (OA), average accuracy (AA), and Kappa
coefficient in the form of mean ± standard deviation to measure the performance of our model.

%\vspace{-0.3cm}
\subsection{Parameters Setting and Experiment Results}
For all experiments, the numbers of labeled samples for training and testing are the same as 3-D CNN experiments\cite{chen2016deep}. The experiment parameters in Indian Pines data as follows: SOMP uses $7\times7$ square window. The window size of 3D-CNN is $31\times31$, and
the convolution kernel size is $4\times4$. For Pixel DAG-RNN, we use 8 neighborhood information and DAG-RNN with 128 dimensions to extract features. In addition, the block size is $13\times13$, the learning rate is 0.005 and dropout is 0.4. The parameters of experiments on University of Pavia data are listed: the window sizes of SOMP and 3-D CNN are $9\times9$ and $27\times27$ respectively. For Pixel DAG-RNN, four $6\times6$ DAGs are used to represent the $11\times11$ image patch. In KSC data, SOMP uses $7\times7$ window, and the parameters of 3-D CNN is the same as University of Pavia data. In Pixel DAG-RNN, the patch size is $13\times13$. An 8 neighborhood UCG to represent the contextual awareness, and then combine four DAGs to approximate the topology of UCG.
Besides 8 neighborhood UCG, we also use a 4 neighborhood UCG to represent an image patch.
The overall results of different methods on three data sets are listed in Table 1.
%\begin{table}[htbp]
%\tiny
%\centering
%\setlength{\abovecaptionskip}{0.2cm}
%\setlength{\belowcaptionskip}{-0.2cm}
%\caption{\footnotesize{OA(\%), AA(\%) and Kappa}}
%\begin{tabular}
%{m{0.3cm}|p{0.4cm}|p{0.9cm}|p{0.9cm}|p{0.9cm}|p{0.9cm}|p{0.9cm}}
%\hline
%\hline
%& method & RBF-SVM &  SOMP & 3D-CNN& 4-P-DAG-RNN& 8-P-DAG-RNN \\
%\cline{1-7}
%\multirow{3}{*}{Pines}&OA&82.20$\pm$0.39 & 95.10$\pm$0.39 & 94.93$\pm$0.65 & 95.13$\pm$0.76 & \textbf{96.42$\pm$0.24}\\
%&AA&87.95$\pm$0.86 & 93.71$\pm$1.33 & 95.37$\pm$1.04 & 95.18$\pm$1.23 & \textbf{96.58$\pm$0.31}\\
%&Kappa&0.863$\pm$0.005 & 0.943$\pm$0.005 & 0.941$\pm$0.008 & 0.944$\pm$0.009 & \textbf{0.959$\pm$0.003}\\
%\cline{1-7}
%\multirow{3}{*}{Pavia}&OA&89.42$\pm$0.40 & 97.34$\pm$0.54 & 98.61$\pm$0.57 & 98.75$\pm$0.29 & \textbf{99.29$\pm$1.75}\\
%&AA&89.62$\pm$0.24 & 95.68$\pm$0.54 & 98.47$\pm$0.40 & 98.28$\pm$0.36 & \textbf{99.07$\pm$0.28}\\
%&Kappa&0.859$\pm$0.005 & 0.964$\pm$0.007 & 0.981$\pm$0.008 & 0.983$\pm$0.004 & \textbf{0.990$\pm$0.002}\\
%\cline{1-7}
%\multirow{3}{*}{KSC}&OA&89.04$\pm$1.35 & 92.65$\pm$1.43 & 94.29$\pm$0.90 & 96.00$\pm$0.52 & \textbf{97.45$\pm$0.72}\\
%&AA&85.61$\pm$1.62 & 92.34$\pm$1.71 & 92.71$\pm$1.35 & 93.13$\pm$0.73 & \textbf{95.75$\pm$1.17}\\
%&Kappa&0.878$\pm$0.015 & 0.918$\pm$0.016 & 0.936$\pm$0.010 & 0.956$\pm$0.006 & \textbf{0.972$\pm$0.008}\\
%\hline
%\hline
%\end{tabular}
%\end{table}

In Table 1, we can find that 8-P-DAG-RNN obtains the best classification performance compared with other methods on OA, AA and Kappa. Compare to other four models which apply spatial-spectral features, RBF-SVM obtain lower classification accuracy because of its only feature extraction on spectral dimension. In University of Pavia and KSC data, 3-D CNN obtains slightly higher accuracies on OA,AA and Kappa. However, in Indian Pines data, SOMP and 3-D CNN obtain nearly the same classification performances due to more categories with fewer training samples which is adverse condition for deep neural network. Because 4-P-DAG-RNN don't consider diagonal connections between units in UCG, it obtains some losses on classification accuracy compared to 8-P-DAG-RNN.
On the whole, our novel model 8-P-DAG-RNN has the best classification performance because of its further use of the spatial contextual dependency.
\begin{table}[htbp]
\tiny
\centering
\setlength{\abovecaptionskip}{0.2cm}
\setlength{\belowcaptionskip}{-0.2cm}
\caption{\footnotesize{OA(\%), AA(\%) and Kappa}}
\begin{tabular}
{m{0.3cm}|p{0.4cm}|p{0.9cm}|p{0.9cm}|p{0.9cm}|p{0.9cm}|p{0.9cm}}
\hline
\hline
& method & RBF-SVM &  SOMP & 3D-CNN& 4-P-DAG-RNN& 8-P-DAG-RNN \\
\cline{1-7}
\multirow{3}{*}{Pines}&OA&82.20$\pm$0.39 & 95.10$\pm$0.39 & 94.93$\pm$0.65 & 95.13$\pm$0.76 & \textbf{96.42$\pm$0.24}\\
&AA&87.95$\pm$0.86 & 93.71$\pm$1.33 & 95.37$\pm$1.04 & 95.18$\pm$1.23 & \textbf{96.58$\pm$0.31}\\
&Kappa&0.863$\pm$0.005 & 0.943$\pm$0.005 & 0.941$\pm$0.008 & 0.944$\pm$0.009 & \textbf{0.959$\pm$0.003}\\
\cline{1-7}
\multirow{3}{*}{Pavia}&OA&89.42$\pm$0.40 & 97.34$\pm$0.54 & 98.61$\pm$0.57 & 98.75$\pm$0.29 & \textbf{99.29$\pm$1.75}\\
&AA&89.62$\pm$0.24 & 95.68$\pm$0.54 & 98.47$\pm$0.40 & 98.28$\pm$0.36 & \textbf{99.07$\pm$0.28}\\
&Kappa&0.859$\pm$0.005 & 0.964$\pm$0.007 & 0.981$\pm$0.008 & 0.983$\pm$0.004 & \textbf{0.990$\pm$0.002}\\
\cline{1-7}
\multirow{3}{*}{KSC}&OA&89.04$\pm$1.35 & 92.65$\pm$1.43 & 94.29$\pm$0.90 & 96.00$\pm$0.52 & \textbf{97.45$\pm$0.72}\\
&AA&85.61$\pm$1.62 & 92.34$\pm$1.71 & 92.71$\pm$1.35 & 93.13$\pm$0.73 & \textbf{95.75$\pm$1.17}\\
&Kappa&0.878$\pm$0.015 & 0.918$\pm$0.016 & 0.936$\pm$0.010 & 0.956$\pm$0.006 & \textbf{0.972$\pm$0.008}\\
\hline
\hline
\end{tabular}
\end{table}

For further expressing contributions of our model on each class, we listed the classification accuracies of all categories on KSC data in Table 2.
From Table 2, we find that 8-P-DAG-RNN obtains the highest accuracy in 8 of 13 categories, such as "Surb", "Slash pine", "Graminoid marsh", "Spartina marsh", "Cattail marsh", "Salt marsh", "Mud flats", "Water". Other five classes obtain the second or third high accuracy because of their fewer training samples. In addition, we analyse the influence of memory length on classification accuracy. As mentioned in 3.2, an image patch can be decomposed into four small image
patches and the size of small image patch is called as memory
length, such as the memory length of Fig. 4 is 4. Through experiments with different memory length(m=\{5,6,7,8\}) of image patches on three datasets, we find that when the memory
length is 6, the University of Pavia dataset can reach the best
performance(OA = 99.29%), and the most suitable memory
length for the Indian Pines and KSC datasets is 7
\begin{table}[htbp]
\tiny
\centering
\caption{\footnotesize{classification accuracy for every class(\%)on the KSC data set.}}
\begin{tabular}{p{1.5cm}|p{0.9cm}|p{0.9cm}|p{0.9cm}|p{0.9cm}|p{0.9cm}}
\hline
\hline
Method & RBF-SVM & SOMP & 3D-CNN & 4-P-DAG-RNN & 8-P-DAG-RNN\\
\hline
Scrub & 92.37$\pm$2.92 & 98.13$\pm$1.31 & 94.48$\pm$2.32 & 97.75$\pm$1.36 & \textbf{98.27$\pm$1.31}\\
Willow swamp & 85.39$\pm$4.08 & \textbf{95.34$\pm$4.88} & 84.08$\pm$10.77 & 90.61$\pm$5.64 & 95.28$\pm$3.80\\
CP hammock & 90.58$\pm$1.91 & \textbf{97.84$\pm$2.14} & 83.30$\pm$5.94 & 93.82$\pm$5.11 & 97.34$\pm$1.34\\
Slash pine & 74.19$\pm$5.71 & 85.20$\pm$4.43 & 83.67$\pm$7.62 & 85.32$\pm$6.74 & \textbf{89.63$\pm$4.61}\\
Oak/Broadleaf & 70.36$\pm$6.08 & 92.33$\pm$6.22 & \textbf{92.86$\pm$3.65} & 73.13$\pm$7.93 & 85.09$\pm$8.29\\
Hardwood & 55.58$\pm$8.61 & \textbf{91.53$\pm$4.29} & 89.87$\pm$4.79 & 84.02$\pm$4.99 & 88.18$\pm$4.81\\
Swamp & 94.34$\pm$3.95 & \textbf{100.00$\pm$0.00} & 99.09$\pm$1.52 & 92.11$\pm$6.14 & 95.30$\pm$4.24\\
Graminoid marsh & 75.09$\pm$7.41 & 78.76$\pm$5.84 & 93.55$\pm$4.70 & 97.24$\pm$2.01 & \textbf{97.73$\pm$1.92}\\
Spartina marsh & 98.27$\pm$1.35 & 94.63$\pm$3.08 & 91.91$\pm$4.66 & 99.83$\pm$0.26 & \textbf{99.95$\pm$0.10}\\
Cattail marsh & 95.31$\pm$2.75 & 95.11$\pm$2.40 & 96.39$\pm$6.34 & 99.54$\pm$0.47 & \textbf{99.69$\pm$0.40}\\
Salt marsh & 95.66$\pm$2.23 & 99.18$\pm$0.52 & 97.48$\pm$2.57 & 99.15$\pm$0.31 & \textbf{99.50$\pm$4.12}\\
Mud flats & 87.68$\pm$5.11 & 72.43$\pm$10.27 & 98.59$\pm$2.42 & 98.11$\pm$1.92 & \textbf{98.78$\pm$1.23}\\
Water & 98.16$\pm$0.50 &\textbf{100.00$\pm$0.00} & \textbf{100.00$\pm$0.00} & \textbf{100.00$\pm$0.00} & \textbf{100.00$\pm$0.00}\\
\hline
\hline
\end{tabular}
\end{table}
%\subsection{Sensitivity Analysis }
%
%\begin{figure}[!htb]
%\centering
%\setlength{\abovecaptionskip}{-0.0cm}
%\setlength{\belowcaptionskip}{-0.4cm}
%\includegraphics[width=1.6in,height=1.2in]{ch/Memory_length.pdf}
%\caption{\footnotesize{The influence of memory length on different data sets.}}
%\label{fig_University}
%\end{figure}
%
%An image patch can be decomposed into four small image patches. We define the size of small image patch as memory length. In Fig.5, a $7\times7$ image patch was decomposed into four $4\times4$ image patches, and the memory length is 4.
%The memory length directly indicates the range of neighborhood information. Therefore, it’s a pivotal parameter in our neural network.
%Fig. 5 shows the overall accuracy (OA) with different memory length (m=\{5,6,7,8\}) of image patches for all the three data sets. We can clearly see that when the memory length is 6, the University of Pavia dataset can reach the best performance(OA = 99.29\%), and the most suitable memory length for the Indian Pines and KSC datasets is 7.
\vspace{-0.3cm}
\section{CONCLUSION}
\label{sec:conclusion}
In this paper, we use Pixel DAG-RNN to extract spectral-spatial features for HSIs classification. It can effectively exploit the spatial correlation of pixels by UCG and then combination of four directed acyclic graphs (DAGs) to approximate the UCG’s topology. In addition, this model also utilizes the advantage of RNN on extracting and using sequence data in network architecture.
The superiority of Pixel DAG-RNN has been verified by experiments on three benchmark HSIs data sets. Further, weights sharing and dropout are used to prevent overfitting. The future work will devote to use pixel information efficiently.

% -------------------------------------------------------------------------
\vspace{-0.4cm}
{\small
\bibliographystyle{IEEEbib}
\addtolength{\itemsep}{-1.5ex} % 缩小参考文献间的垂直间距
\bibliography{refs}
}

\end{document}